\documentclass[9pt,twocolumn,twoside]{template} 
\templatetype{twocolumns} 
\title{ChatGPT-4 Outperforms Experts and Crowd Workers in Annotating Political Twitter Messages with Zero-Shot Learning}
\author[a,c,1]{Petter Törnberg}
\affil[a]{Amsterdam Institute for Social Science Research (AISSR), University of Amsterdam}
\leadauthor{Törnberg} 
\authordeclaration{The author has no competing interests to declare.}
\correspondingauthor{\textsuperscript{2}To whom correspondence should be addressed. E-mail: p.tornberg@uva.nl}

\begin{abstract}
This paper assesses the accuracy, reliability and bias of the Large Language Model (LLM) ChatGPT-4 on the text analysis task of classifying the political affiliation of a Twitter poster based on the content of a tweet. The LLM is compared to manual annotation by both expert classifiers and crowd workers, generally considered the gold standard for such tasks. We use Twitter messages from United States politicians during the 2020 election, providing a ground truth against which to measure accuracy. The paper finds that ChatGPT-4 has achieves higher accuracy, higher reliability, and equal or lower bias than the human classifiers. The LLM is able to correctly annotate messages that require reasoning on the basis of contextual knowledge, and inferences around the author's intentions -- traditionally seen as uniquely human abilities. These findings suggest that LLM will have substantial impact on the use of textual data in the social sciences, by enabling interpretive research at a scale. 
\end{abstract}

\dates{This manuscript was compiled on \today}

\begin{document}

\maketitle
\thispagestyle{firststyle}
\ifthenelse{\boolean{shortarticle}}{\ifthenelse{\boolean{singlecolumn}}{\abscontentformatted}{\abscontent}}{}

\dropcap{T}exts are culturally and socially situated within the ideas, values and beliefs about the world within which humans operate. Their interpretation thus requires deep contextual knowledge and the ability to ``put ourselves in the shoes of others'' -- seen as distinctly human resources and capacities. While interpretation lies at the very heart of the social sciences \citep[28]{weber_sociological_1994}, it has thus been seen as almost intrinsically qualitative; ``meaning has to be understood, it cannot be measured or counted, and hence there is always an interpretive or hermeneutic element in social science'' \citep[17]{sayer1999realism}. 

Researchers seeking to use textual data have thus had to pursue to one of two strategies for analyzing textual data: so-called expert coders (such as colleagues, students, or research assistants), or crowd-workers on online labor platforms, in particular Amazon's MTurk. These strategies are often combined, for instance by using trained annotators to create a smaller dataset, which can be used to evaluate the quality of crowd-workers that are then employed to expand the volume of data. But while humans have been considered the unrivaled gold standard for interpretive tasks \citep{murthy_twitsenti_2019,van_atteveldt_validity_2021}, we bring several limitations: we are slow, costly, biased, and have limited attention-spans, hence restricting interpretive research to relatively small-N studies – in turn leading to complaints of lacking rigor and replicability, and limiting the potential in Big Data research. As a result, conventional forms of interpretive textual research thus tends to miss patterns of language use are ``not directly observable, because they are realized across thousands or millions of words of running text, and because they are not categorical but probabilistic'' \citep[204]{stubbs_grammar_1994}.  While computational methods for analyzing textual data -- such Natural Language Processing and Machine Learning -- have evolved quickly in recent years \citep{eisenstein2019introduction}, common methods that build on bag-of-words, lexical meanings and sentence semantics tend to work poorly for tasks requiring complex inferences on the basis of knowledge about the world or assumptions of the author's mental state and intentions.

The recent rise of generative AI may however be at the cusp of changing this. In recent months, ChatGPT has become a global sensation, and one of the quickest growing consumer products of all time. ChatGPT is a pre-trained model based on a massive neural network with billions of parameters, and trained on hundreds of billions of words of text from the Internet and digitized books \citep{brown2020language}. While skeptics have dismissed these forms of Large Language Models (LLM) as merely ``fancy autocomplete'', the models have demonstrated several surprising emergent capabilities \citep{wei_emergent_2022} – even showing ``sparks of general intelligence'' \citep{bubeck_sparks_2023} -- including the capacity for few-shot and zero-shot classification tasks \citep{kojima_large_2022,brown2020language}. As these models are general rather than field or task-specific, they appear to capture some of the context needed for interpretation – enabling it to even surpass fine-tuned models. Recent studies have found that LLMs perform well for a wide range of purposes, including ideological scaling \citep{wu_large_2023}, text annotation tasks \citep{gilardi_chatgpt_2023}, and for simulating samples for survey research \citep{argyle_out_2022}.

This paper explores the potential of LLMs for text analysis tasks whose interpretation require reasoning on the basis of contextual knowledge. Previous studies have been limited by lacking ground truth with which to compare accuracy, or comparing LLM's only with crowd workers, known to be unreliable in terms of accuracy and data quality \citep{chmielewski2020mturk}. We move beyond existing research by focusing on the recently released ChatGPT-4 model, assessing its accuracy compared to both MTurk workers and experts, on a realistic social scientific task for which ground truth is available. Focusing on identifying the political party affiliation of Twitter posters, we find that ChatGPT-4 reliably outperform both crowd-workers and expert coders. We discuss the implications, risks, and limitations of this potential paradigm shift for using text-as-data, as the large-N interpretive works hits at the core of the traditional division between the quantitative and qualitative paradigms.

\newpage

\section*{ChatGPT and the rise of Generative AI}
ChatGPT is an AI chatbot developed by OpenAI and introduced in November 2022. Given a prompt as an initial text input, ChatGPT generates a response, simulating a conversation with the user. It is a member of the generative pre-trained transformer (GPT) family of language models and is based on OpenAI's LLMs GPT-3.5 and GPT-4 \citep{brown2020language}. These models are built on highly complex artificial neural networks trained to predict the next token given a set of existing tokens. The GPT models were trained on a vast corpus of text, including a significant share of the Internet and all books ever written. The ChatGPT model was furthermore fine-tuned using both supervised and reinforcement learning techniques to achieve human-like responses \citep{ziegler_fine-tuning_2019}. The supervised learning consisted of human trainers playing both the user and the AI assistant in simulated conversations. In the reinforcement learning step, the model generated text that was then ranked by human annotators from most preferred to least preferred, allowing the LLM to further improve its responses \citep{schulman_chatgpt_2022}.

While this type of transformer-based language models originally functioned as a form of sophisticated autocomplete, they began taking on surprising emergent properties when they reach a certain size. In line with Anderson's \cite{anderson_more_1972} famous suggestion that ``more is different'', the models appear to go through a form of phase transition, bringing about new capacities for which they were not explicitly trained \citep{wei_emergent_2022}. For instance, ChatGPT appears to compose information, as opposed to merely regurgitating it. It is able to operate in and translate between several languages. It is able to generate computer code based on instruction in natural language, and to write prose or poetry on any topic in a given style. For the purposes of this paper, the most relevant such emergent capacity is the ability to engage in what appears as a form of reasoning on the basis of known contextual information, allowing the model to interpret textual statement with zero- or few-shot learning. 

However, while ChatGPT displays several noteworthy capacities and have drawn substantial international attention, the model still has several important limitations. For instance, when facing questions that the model does not ``know'' the answer to, the model is prone to ``hallucinate'' \citep{bang2023multitask}, that is, it produces confident-sounding ``bullshit'' \citep{frankfurt2005bullshit} answers. The model is furthermore limited when it comes to symbolic logic -- failing even simple mathematical tasks when faced with large-enough numbers. More seriously, the models reproduces the biases and racism embodied in the data on which it was fed, and LLMs such as ChatGPT are therefore prone to generate text involving negative stereotypes, for instance in relation to race and gender \citep{lucy_gender_2021}. To address this, OpenAI has imposed several ``guardrails'' to prevent the chatbot from expressing offensive views. 

Social scientists have recently begun studying the possibilities of using LLMs such as ChatGPT for social scientific research. Ornstein et al. \citep{ornstein_how_2022} found that GPT-3 can be effectively applied in few-shot learning contexts to text-as-data tasks in political science, including sentiment analysis, ideological scaling, and topic modeling. Argyle et al. \citep{argyle_out_2022} found that biases in GPT-3 are fine-grained and demographically correlated, and can be used to simulate partisan responses from a wide variety of human communities. Palmer and Spirling \citep{palmer_large_2023} found that LLMs can argue in persuasive and novel ways about politics. Wu et al \citep{wu_large_2023} found that ChatGPT can be used to create a scale of politicians' ideologies through pair-by-pair comparison. In this paper, we will assess the accuracy of ChatGPT-4 for annotating and classifying political messages on Twitter.

\section*{Method: Using ChatGPT to classify Twitter messages}
We use data collected by van Vliet et al \citep{van_vliet_twitter_2020}, selecting all messages posted to Twitter from all United States senators during the two months preceding the US 2020 election, that is, between September 3rd, 2020, and November 3rd, 2020. We use messages from politicians as these offer a ground truth: we know their political affiliations and can therefore objectively evaluate the accuracy of the ChatGPT answers, and compare with the accuracy of experts. However, the method could be similarly applied to examine the political views or positions of normal social media users. From this dataset, retweets and replies were removed, as were tweets containing an URL, as were messages shorter than 100 characters (as these were not long enough to contain political content.) Random samples of 250 tweets from Republican and from Democratic politicians were selected, resulting in a balanced sample with 500 tweets.

The resulting set of tweets contain a mix between expressions of view on particular policy positions, comments on recent political events (such as the COVID-19 Pandemic, or the nomination of Amy Coney Barrett to the Supreme Court), self-congratulatory messages for some policy victory, calls for campaign contributions, and the occasional personal news. Some of the tweets can be easily classified as being from a Democrat or Republican -- such as laudatory references to the then-sitting President Donald Trump, or attacks against the opposing party. Other require detailed knowledge of US politics, such as the positions of the two parties on particular political issues in Oklahoma. Other require the capacity to consider the author's intention in relation to the electoral considerations of the two parties, such as Bible quotes, speaking to the Evangelical base of the Republican party. Finally, some tweets are nearly impossible to guess, such as the lamenting of the death of a friend. 

We used ChatGPT-4 through the API to classify each message, using the following instruction:

\begin{quote}
``You will be given a set of Twitter posts from different US politicians, sent during the two months preceding the 2020 US presidential election, that is, between September 3rd, 2020, and November 3rd, 2020. Your task is to use your knowledge of US politics to make an educated guess on whether the poster is a Democrat or Republican. Respond either `Democrat' or `Republican'. If the message does not have enough information for an educated guess, just make your best guess.''
\end{quote}

The API was then given the Twitter messages, selected in random order. The model was run with high and low values for temperature -- a parameter that controls the level of randomness or ``creativity'' in the generated text.  Since the responses are stochastic, we ran the model 5 times at a low temperature (0.2), and at a high temperature (1.0) to capture variability in responses. The model was thus run for a total of 5000 times.

To assess the accuracy, reliability and bias of ChatGPT with the current gold standard for textual annotation, we compare the resulting data with two manual datasets: MTurk classification, and expert classifiers. 

MTurk is an online platform for click-labor, which gives each task to a large number of workers, and to thus use the ``wisdom of crowd'' effect (a group generally has higher accuracy than any given individual) to reach high levels of accuracy. The MTurk workers were given the same instructions as ChatGPT. 

MTurk classification vary significantly in quality depending on the details of how the tasks are implemented, making it perilous to compare models with MTurk accuracy without additional verification \citep{chmielewski2020mturk}.  
In the pre-study testing, MTurk was run without stringent quality evaluation, resulting in a mean accuracy that was not significantly different from random. To maximize accuracy of the MTurk predictions, best-practices were thus used to ensure high quality responses from the MTurk crowd workers: only Master Qualified workers (a special group of elite MTurk workers whose abilities have been verified) residing in the United States were allowed to respond. The workers were given the questions in batches of 50, each with 4 control questions in which the author's political affiliation was evident (e.g., ``If one thing is certain, I am a Democrat!''). If the worker failed any of these questions, their responses were automatically rejected. Even when passing these tests, the work was rejected if the answer accuracy was not statistically different from random. Each question was answered by 10 independent crowd-workers. 

To verify the quality of the MTurk response, two political science researchers also manually classified all messages. While these scholars do not specifically work on US politics, they have a strong general understanding, and thus provide a baseline of accuracy. These experts were given the 500 messages in random order, and again the same instructions as ChatGPT. 

\section*{Results}

\begin{figure}
\centering
\includegraphics[width=1\linewidth]{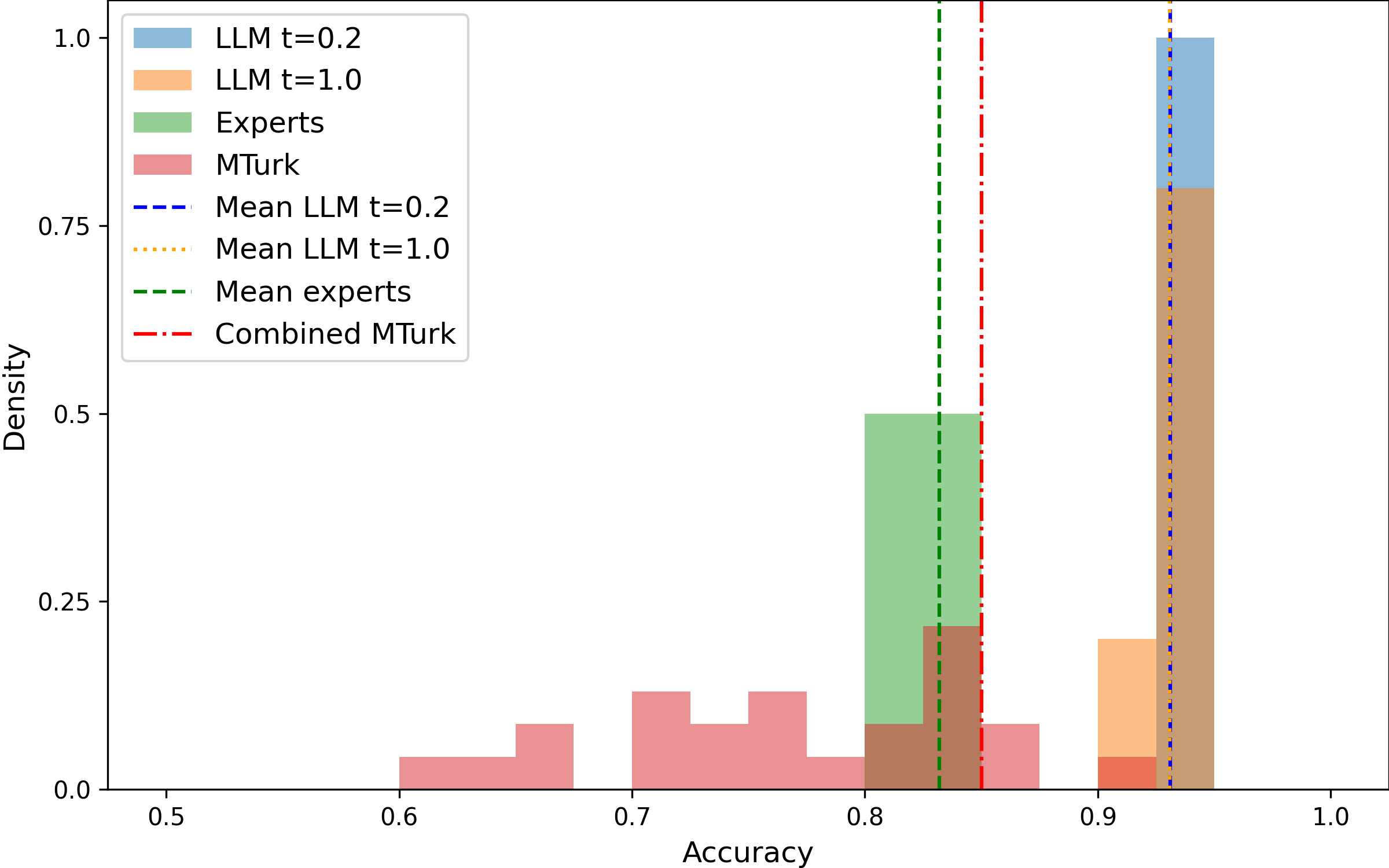}
\caption{A normalized density histogram of the response accuracy by crowd worker, experts, and the LLM models. The dashes lines show the mean response accuracy for LLMs and experts. The combined MTurk line shows the accuracy of the majority response for each question, as this is how crowd-worker answers tend to be employed. As can be seen, the LLMs outperform all individual human classifiers. The mean accuracy lines for the two temperatures of the LLM nearly precisely overlap.}
\label{fig:accuracy}
\end{figure}

Figure \ref{fig:accuracy} shows the distribution of accuracies per worker, model run, and expert, as well as the average accuracies. As the figure shows, the LLM outperforms all individual human classifiers, as well as the combined crowd workers. For MTurk, the accuracy of the plurality response of the 10 workers are reported, as the ``wisdom of crowds'' effect means that the accuracy of the combined answers is often higher than any individual. As the figure shows, the expert classifiers have higher accuracy than most individual  MTurk classifiers, but lower than their combined response. The LLM has a slightly higher accuracy and lower spread at lower temperatures.

\begin{figure}
\centering
\includegraphics[width=.8\linewidth]{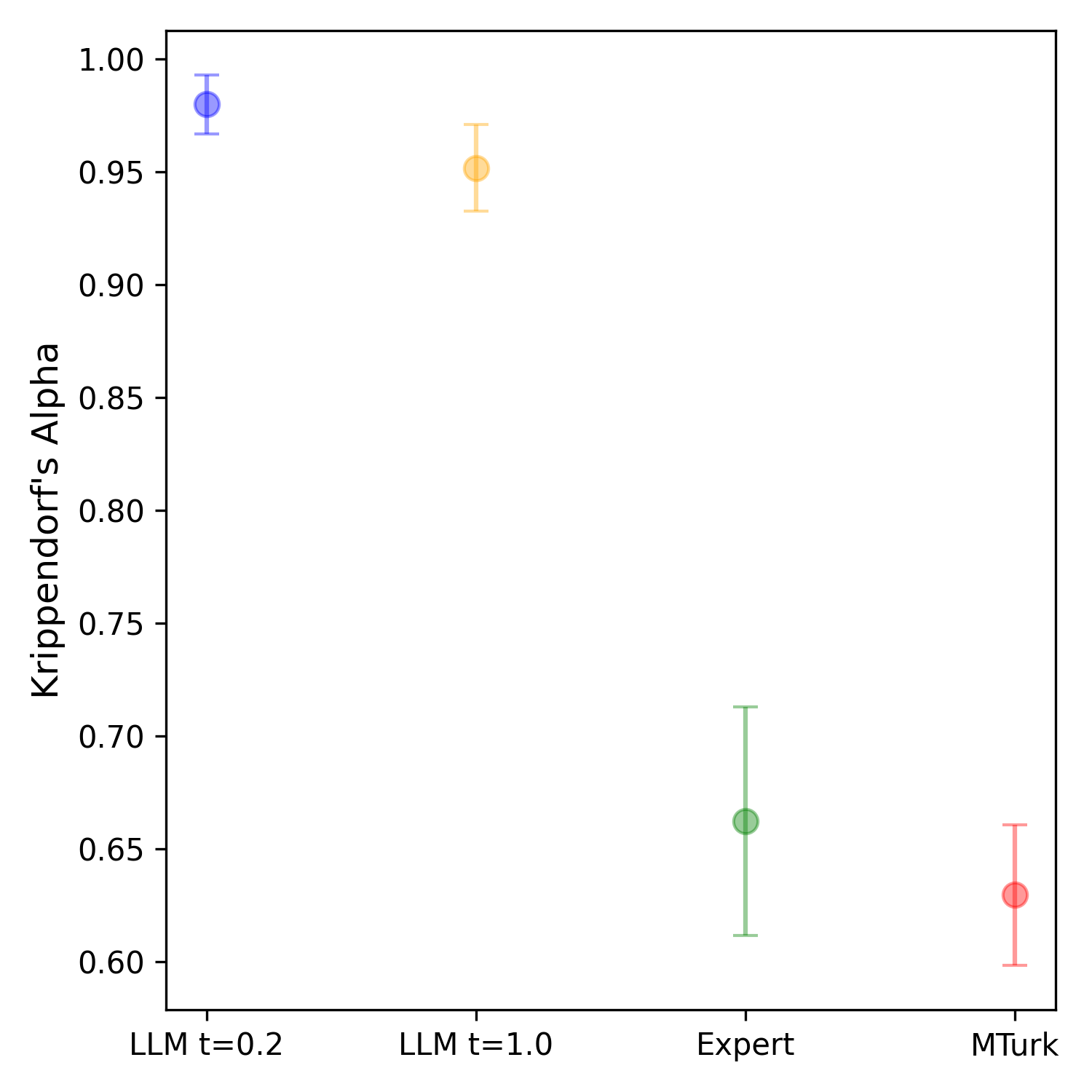}
\caption{The Krippendorf's Alpha measure of intercoder reliability, with 95\% bootstrap confidence interval. The LLM offers substantially higher levels of reliability than the human coders.}
\label{fig:krippen}
\end{figure}

Reiss \citep{reiss_testing_2023} suggests that ChatGPT may suffer from limited reliability, giving stochastic answers and thereby limiting the usefulness for annotation and classification tasks. However, this study does not compare the reliability with human classifiers. To measure intercoder reliability, we calculate Krippendorf's Alpha. Figure \ref{fig:krippen} shows the result. We here see that compared to human classifiers, the LLM in fact has a much higher reliability for the task at hand. As expected, the LLM's reliability is higher at lower temperature. 

\begin{figure}
\centering
\includegraphics[width=.8\linewidth]{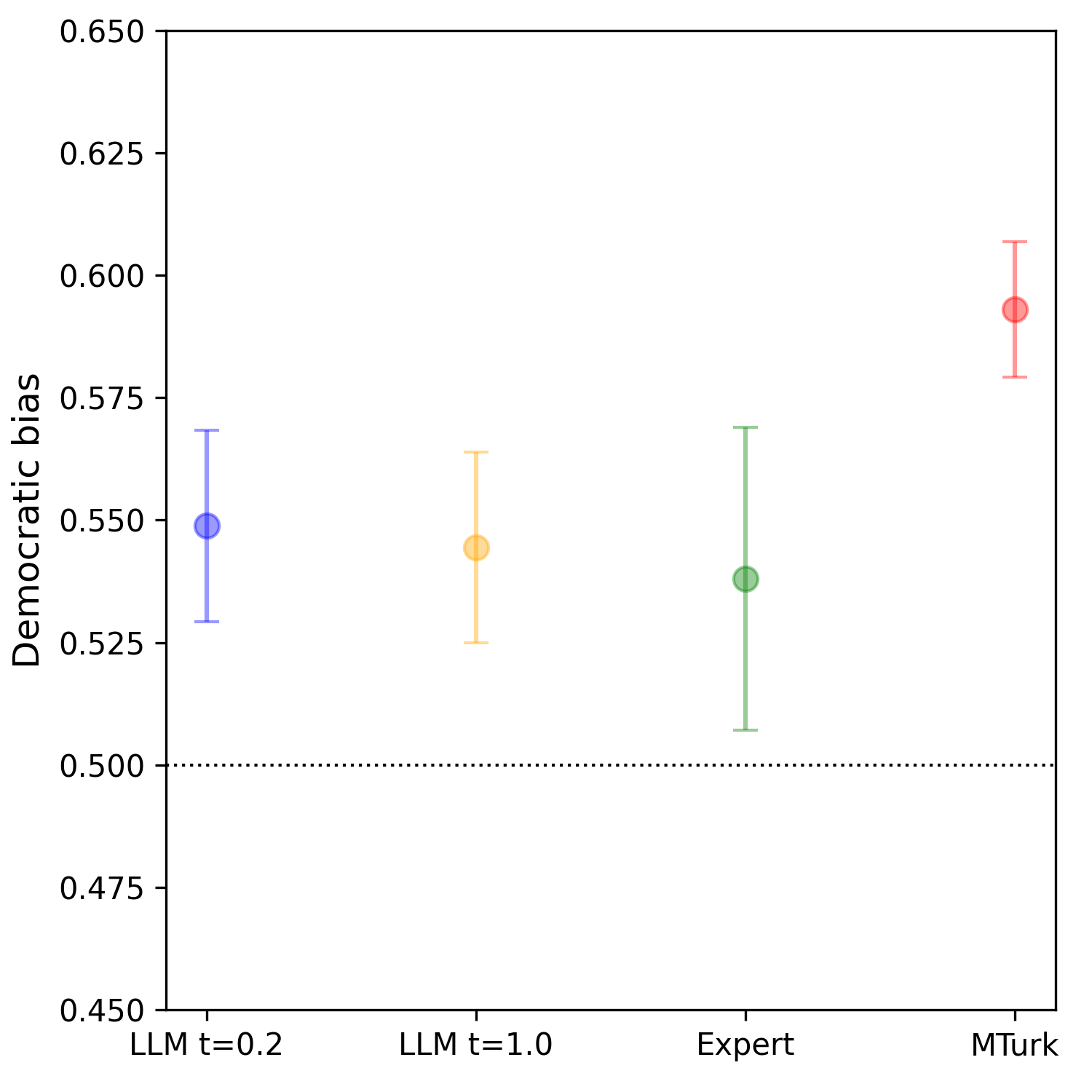}
\caption{All coders are biased to guessing Democrat over Republican. The LLMs and experts are similar in the level of bias, while the MTurk classifiers have a significantly stronger bias.}
\label{fig:bias}
\end{figure}

To examine whether there is any bias in the model compared to the manual classification in terms of whether the model is more likely to respond with one party over the other, we examine the fraction of responses for each party for LLM, MTurk and the expert annotators. Figure \ref{fig:bias} shows that all respondents are significantly biased toward guessing that the tweet author is a Democrat, including the LLM. The biases of the expert classifiers and the LLM are not significantly different, while the MTurk classifiers are significantly more biased. 

Finally, we examine select cases of how ChatGPT responded in relation to tweets that require particular interpretive skills, by following up the question with a request for ChatGPT to ``motivate your response''. 
First, a tweet whose meaning is implicit and depends on contextual information: ``A friend, a neighbor, a mom, who could soon serve on the highest court in the land''. ChatGPT correctly responded Republican, as did 7 of the MTurk workers. ChatGPT motivates its answer with: 

\begin{quote}
``This tweet is likely referring to the nomination of Amy Coney Barrett to the Supreme Court, which took place during the specified time frame. Barrett was nominated by then-President Donald Trump, a Republican, and her nomination was supported by the Republican party. The positive tone of the tweet suggests that the poster is in favor of her nomination, which leads me to believe that the poster is a Republican.''
\end{quote}

Second, a tweet containing a Bible quote (``good indeed is the LORD, His mercy endures forever, his faithfulness lasts through every generation. Psalms 100:5''). Interpreting this requires understanding that the intent of the post is to emphasize values primarily associated Republican voters. The LLM and 9 out of 10 MTurk workers correctly guessed that the author was Republican.

\begin{quote}
``I guessed that the poster is a Republican because the tweet includes a quote from the Bible, specifically Psalms 100:5. While both Democrats and Republicans can be religious, Republicans are generally more likely to emphasize their religious beliefs and values in public statements, as they tend to have a stronger connection with conservative Christian voters.''    
\end{quote}

While the stringency of the reasoning can be questioned and it is not known what processes are in fact underlying the responses of the model, the reasoning does appear plausible and convincing at face value and the accuracy of the model's guesses are undeniable.

\section*{Discussion}
This brief paper has assessed the accuracy, reliability and bias of ChatGPT-4 for a zero-shot learning classification task of identifying political affiliations based on Twitter messages. The findings suggest that LLMs may already outperform the standard approaches of crowd-workers and expert classifiers, offering higher accuracy, higher reliability, and lower or equal levels of bias -- even for task that require reasoning on the basis of contextual knowledge and assumptions of the author's intentions.

The possibility of using LLMs for such tasks have significant scientific implications, enabling large-scale and replicable interpretative research with low costs and barriers of entry -- thus further blurring the traditional lines between quantitative and qualitative approaches. LLMs do not only offer higher data quality than conventional means of text analysis, but are also orders of magnitude less costly and time-consuming. The models can furthermore be used with limited technical skills, with much the same instructions as would be used for human coders.

However, several caveats must be made. The capacity of LLMs to do ``zero'' or ``few-shot'' learning is an emergent property, for which the models are not explicitly trained. How LLMs acquire such capabilities is not well understood at a conceptual level, and we therefore do not fully know their scope or limitations. Second, LLMs can be sensitive to researcher choices, such as prompt design. Like human coders, LLMs are prone to misunderstanding poorly formulated tasks, but unlike human coders, they are also unlikely to admit uncertainty or inform the researcher when they struggle to comprehend their task. Identifying the optimal instructions may thus require a careful iterative process of experimentation and validation. Third, LLMs embody the biases and prejudices of the texts on which they are trained. ChatGPT in particular has been found to display problematic gender and racial stereotypes, when users have been able to bypass the imposed guardrails. It remains poorly understood if and how such biases affect the models' performance on specific analysis tasks. The solution to all of these issues is careful validation of the model's performance on specific tasks -- and never to assume that the approach will work ``out of the box'' \citep{grimmer2013text}. While training may no longer be necessary for many tasks, validation remains crucial, to make sure that the models are measuring what we intend -- and without problematic biases.

The pace of innovation of LLMs is astounding, and while several fundamental issues remain, it seems likely that LLMs will fundamentally transform social scientific research by opening the possibility of large-N interpretive research -- representing a new paradigm of text-as-data research. At the same time, such advances raise important epistemological challenges, as these models are black-boxed and their capacities are emergent and poorly understood. Beyond continued technical exploration of the capacities of these models and the development of standards for rigorous research, a key task for researchers going forward will be to tackle their profound epistemological implications for social scientific research.

\acknow{The author would like to acknowledge the anonymous crowd workers and expert classifiers for annotating the tweets, and Dr. Juliana Chueri for her suggestions, input and support.}

\showacknow{} 

\subsection*{References}
\bibliography{bibs}

\end{document}